\documentclass[final,5pt]{elsarticle}

\usepackage{mathrsfs}

\usepackage{epsfig}
\usepackage[centertags]{amsmath}
\usepackage{graphics}
\usepackage{epstopdf}
\usepackage{subfigure}
\usepackage{booktabs}
\usepackage{amsthm}
\usepackage{amsmath}
\usepackage{amssymb}
\usepackage{psfrag}
\usepackage{color}

\usepackage[ruled]{algorithm2e}
\usepackage{algorithmic}
\usepackage{setspace}

\DeclareMathOperator*{\argmin}{arg\,min}

\DeclareMathOperator*{\Min}{minimize}

\newcommand{\reals}{{\mathbb{R}}}

\newcommand{\sgn}{{\mathrm{sgn}}}

\newtheorem{theorem}{Theorem}[section]
\newtheorem{lemma}{Lemma}[section]
\newtheorem{remark}{Remark}[section]

\newtheorem{assumption}{Assumption}[section]

\newcommand{\vx}{{\bf x}}
\newcommand{\vt}{{\bf t}}

\newcommand{\vu}{{\bf u}}
\newcommand{\vv}{{\bf v}}

\newcommand{\vzero}{{\bf 0}}
\newcommand{\vw}{{\bf w}}
\newcommand{\vy}{{\bf y}}

\newcommand{\vU}{{\bf U}}

\def\cL{{\cal L}}

\newproof{pf}{Proof}

\begin{document}
%\doi{10.1080/0020717YYxxxxxxxx}
%\issn{1366-5820}
%\issnp{0020-7179}
%\jvol{00}
%\jnum{00}
%\jyear{2010}
%\jmonth{January}

\markboth{X. Huang and M. Yan}{Signal Processing}

%\articletype{Research Article}

\title{Nonconvex Penalties with Analytical Solutions for One-bit Compressive Sensing}

%\tnotetext[fn1]{ }

\author[sjtu]{Xiaolin Huang}
\ead{xiaolinhuang@sjtu.edu.cn}
\author[msu]{Ming Yan\corref{cor1}}
\ead{yanm@math.msu.edu}
\address[sjtu]{Institute of Image Processing and Pattern Recognition, Shanghai Jiao Tong University, and MOE Key Laboratory of System
Control and Information Processing, Shanghai 200240, P.R. China.}
\address[msu]{Department of Computational Mathematics, Science and Engineering and Department of Mathematics, Michigan State University, East Lansing, MI 48824, USA}
%\address{}

\cortext[cor1]{Corresponding author.}

\begin{abstract}
One-bit measurements widely exist in the real world and can be used to recover sparse signals. This task is known as one-bit compressive sensing (1bit-CS). In this paper, we propose novel algorithms based on both convex and nonconvex sparsity-inducing penalties for robust 1bit-CS.
{We consider the dual problem, which has only one variable and provides a sufficient condition to verify whether a solution is globally optimal or not.
For positive homogeneous penalties, a globally optimal solution can be obtained in two steps: a proximal operator and a normalization step.
For other penalties, we solve the dual problem, and it needs to evaluate the proximal operators for many times.
Then we provide fast algorithms for finding analytical solutions for three penalties: minimax concave penalty (MCP), $\ell_0$ norm, and sorted $\ell_1$ penalty. }
Specifically, our algorithm is more than $200$ times faster than the existing algorithm for MCP. Its efficiency is comparable to the algorithm for the $\ell_1$ penalty in time, while its performance is much better than $\ell_1$. Among these penalties, sorted $\ell_1$ is most robust to noise in different settings.
\end{abstract}


\maketitle

\section{Introduction}

Analog-to-digital converting (ADC) is a necessary process in digital processing, and the choice of the bit-depth is an important issue. The extreme case is to use one-bit measurements, which enjoy many advantages, e.g., they can be implemented by one low power comparator running at a high rate. Mathematically, one-bit compressive sensing (1bit-CS) is to recover a $K$-sparse vector $\vx \in \reals^n$ ($\|\vx\|_0\leq K$) from $m$ one-bit quantized measurements
\begin{equation}\label{noise-data}
y_i=\sgn(\vu_i^\top\vx+\varepsilon_i),
\end{equation}
where $\vu_i \in \reals^n$ is the $i$th sensing vector, $\varepsilon_i$ is the noise in the measurement, and the function $\sgn$ returns $1$ for a positive number and $-1$ otherwise. The sensing system and measurements are represented by $\vU = [\vu_1, \vu_2, \ldots, \vu_m]$ and $\vy = [y_1, y_2, \ldots, y_m]^\top$, respectively. {Due to the low power and high sampling rate, one-bit measurements have been applied in the estimation of frequency, phase, and direction of arrival (DOA)~\cite{host1995lower, host2000effects, bar2002doa}. For example, in the DOA estimation, a radar with one-bit measurements has a higher scan speed than others. One-bit measurements are also attractive in distributed networks \cite{chen2010performance, chen2010nonparametric}, where the use of one-bit measurements largely reduces the communication load.}

%The task is essentially to reconstruct a sparse signal from one-bit measurements and is called One-bit Compressive Sensing (1bit-CS).
{If the underlying signal is sparse, then sparsity pursuit techniques can help signal recovery, which is similar to the regular compressive sensing. Therefore},
since its proposal by~\cite{boufounos20081bit}, 1bit-CS has attracted much attention in both the signal processing society (\cite{laska2011trust, yan2012robust, plan2013robust, baraniuk2017exponential}) and the machine learning society (\cite{zhang2014efficient, zhu2015towards, chen_onebit_2015, awasthi2016learning}). Because the one-bit information has no capability to specify the magnitude of the original signal, we assume $\|\vx\|_{2} = 1$ without loss of generality (there is also some work on norm estimation, see, e.g., \cite{knudson2016one}), and 1bit-CS can be explained as finding the sparest vector on the unit sphere that coincides with the observed signs, i.e,
\begin{eqnarray}\label{1bit}
\begin{array}{rl}
\Min\limits_{\vx \in \reals^n} & \|\vx\|_{0}, \\
\mathrm{subject~to} & y_i =\sgn(\vu_i^\top \vx), ~~  \forall i = 1, 2, \ldots, m,\\
 & \|\vx\|_{2} = 1.
\end{array}
\end{eqnarray}
This is an NP-hard problem, and several algorithms are developed to approximately solve it or its variants~\cite{boufounos20081bit,laska2011trust,boufounos2009greedy,xu2014bayesian}.
The constraint in~\eqref{1bit} does not tolerate noise or sign flips, and it may exclude the real signal from the feasible set. Additionally, the feasible set may be empty, and there is no solution for~\eqref{1bit}.
One way to deal with noise and sign flips is to replace the constraint $y_i=\sgn(\vu_i^\top \vx)$ by a loss function.
For example, the one-sided $\ell_1$ loss and the one-sided $\ell_2$ loss are considered in~\cite{jacques2013robust} and~\cite{yan2012robust}; the linear loss is used in~\cite{plan2013robust} and~\cite{zhang2014efficient}.
It is reported that the linear loss generally outperforms the one-sided $\ell_1$/$\ell_2$ loss. Moreover, with proper regularization terms and constraints, the linear loss minimization can be solved analytically and enjoys great computational effectiveness.

In regular CS problems, nonconvex penalties have been insightfully investigated and widely applied to enhance sparsity.
Similarly, those nonconvex techniques are applicable to 1bit-CS, and the recovery performance is expected to be improved.
One obvious barrier is that nonconvex penalties lead to nonconvex problems, which are usually difficult to solve.
An interesting result is recently represented in~\cite{zhu2015towards}, which gives analytical solutions for two nonconvex penalties, namely the smoothly clipped absolute deviation (SCAD,~\cite{fan2001variable}) and minimax concave penalty (MCP,~\cite{zhang2010nearly}).
Also~\cite{chen_onebit_2015} proposes an algorithm for 1bit-CS using the $k$-support norm.
These nonconvex penalties are shown to obtain better results than convex ones in both theory and practice~\cite{zhu2015towards,chen_onebit_2015} and, therefore, have been extended to other applications including the multi-label learning task~\cite{qiu2017nonconvex}.

In this paper, we discuss more convex and nonconvex penalties, for which analytical solutions can be obtained, and we provide fast algorithms for finding these solutions.
These penalties include SCAD, MCP, $\ell_p$-norm ($0 \leq p \leq +\infty$,~\cite{chartrand2008restricted}), $\ell_1$-$\ell_2$ norm~\cite{yin2015minimization}, sorted $\ell_1$ penalty~\cite{huang2014two, bogdan2013statistical}, and so on.
%In particular, our algorithm is 200 times faster than the algorithm proposed in~\cite{zhu2015towards} for MCP.
%With these efficient algorithms, we compare and evaluate the performance of different nonconvex penalties via numerical experiments.
The contributions of this paper can be summarized as follows.
\begin{itemize}
\item We analyze a generic model for 1bit-CS and provide a sufficient condition for the global optimality.
\item For positive homogeneous penalties, we show that an optimal solution can be obtained in two steps: a proximal operator and a normalization step.
For general penalties, we provide a generic algorithm by solving the dual problem.
\item We provide algorithms for finding analytical solutions for three nonconvex penalties: MCP, $\ell_0$ norm, and the sorted $\ell_1$ penalty.
These algorithms are much faster than the existing 1bit-CS algorithms for nonconvex penalties and even comparable to that for the convex $\ell_1$ minimization problem, e.g., our algorithm is averagely 200 times faster than the algorithm given in~\cite{zhu2015towards} for MCP.
In addition, we compare these nonconvex penalties with the convex $\ell_1$ penalty and show that the sorted $\ell_1$ performs the best in both performance and computational time.
%\item With this approach, we are able to analyze theoretical properties such as sampling complexity for nonconvex penalties, which is beyond the focus of this paper.
\end{itemize}

The rest of this paper is organized as follows.
Section~\ref{sec:related} briefly reviews the existing related 1bit-CS algorithms.
The main contributions, i.e., analytical solutions for different penalties and corresponding algorithms, are presented in Section~\ref{sec:opt}.
The numerical experiments are reported in Section~\ref{sec:num}.
We end this paper with a brief conclusion.

\section{Related Works}\label{sec:related}
Model~\eqref{1bit} for 1bit-CS has two main disadvantages: i) it is difficult to solve because of the $\ell_0$ norm in the objective and the constraint $\|\vx\|_2=1$; ii) the constraint $y_i=\sgn(\vu_i^\top\vx)$ does not consider noisy sign measurements.

Several approaches are given to deal with both disadvantages.
For the nonconvexity, the $\ell_0$ norm is replaced by the $\ell_1$ norm, and the constraint $\|\vx\|_2=1$ is replaced by other convex constraints.
The first convex model \cite{plan2013one} for 1bit-CS is
\begin{eqnarray}\label{1bit-l1-convex}
\begin{array}{rl}
\Min\limits_{\vx \in \reals^n} &  \|\vx\|_{1},  \\
\mathrm{subject~to} &  y_i (\vu_i^\top \vx) \geq 0, ~ \forall i = 1, 2, \ldots, m,\\
 & \|\vU^\top \vx\|_{1} = r,
\end{array}
\end{eqnarray}
where $r$ is a given positive constant. In fact, the solutions for all positive $r$'s have the same direction and the difference is only on the magnitudes of the reconstructed signals.

However,~\eqref{1bit-l1-convex} still can not be applied when there are noisy measurements, because, it, same as~\eqref{1bit}, requires the sign consistence in the measurements.
Noisy measurements come from both the noise during the acquisition before the quantization  and sign flips during the transmission.
%Those will make $y_i (\vu_i^\top \vx) \geq 0, \forall i$ infeasible, or feasible but the true signal is excluded from the feasible set.
To deal with noisy measurements, \cite{jacques2013robust} introduces the following robust model using the one-sided $\ell_1$ norm,
\begin{eqnarray*}%\label{BIHT1}
\begin{array}{rl}
\Min\limits_{\vx\in\reals^n}    &\displaystyle  \frac{1}{m} \sum_{i=1}^m \max \left\{0, -y_i (\vu_i^\top \vx)\right\},\\
\mathrm{subject~to} & \|\vx\|_{2} = 1, \\
              & \|\vx\|_{0} = K.
\end{array}
\end{eqnarray*}
The robust model using the one-sided $\ell_2$ norm is also introduced.
Several modifications are designed by~\cite{yan2012robust}, \cite{bahmani2013robust}, and \cite{dai2014noisy} to improve their robustness to sign flips and noise.

The linear loss for robost 1bit-CS attracts more attention because of its good performance and simplicity.
Based on the linear loss, many results on sampling complexities are given recently~\cite{plan2013robust,zhang2014efficient,zhu2015towards,chen_onebit_2015}.
In \cite{plan2013robust}, the first model using the linear loss for 1bit-CS is proposed and takes the following formulation,
\begin{eqnarray}\label{Plan}
\begin{array}{rl}
\Min\limits_{\vx\in\reals^n}     &\displaystyle - \frac{1}{m} \sum_{i=1}^m y_i (\vu_i^\top \vx),\\
\mathrm{subject~to} &  \|\vx\|_{2} \leq 1, \\
              &  \|\vx\|_{1} \leq s,
\end{array}
\end{eqnarray}
where $s$ is a given positive constant.
One can also put the $\ell_1$-norm in the objective instead of in the constraint, resulting in the problem  given by~\cite{zhang2014efficient},
\begin{eqnarray}\label{Zhang}
\begin{array}{rl}
\Min\limits_{\vx\in\reals^n}    &\displaystyle \mu \|\vx\|_{1} - \frac{1}{m} \sum_{i=1}^m y_i (\vu_i^\top \vx),\\
\mathrm{subject~to} & \|\vx\|_{2} \leq 1,
\end{array}
\end{eqnarray}
where $\mu$ is the regularization parameter for the $\ell_1$-norm. Note that the unit sphere constraint $\|\vx\|_{2} = 1$ is relaxed to the unit ball constraint $\|\vx\|_2\leq 1$ in~\eqref{Plan} and~\eqref{Zhang}. As illustrated by~\cite{zhang2014efficient}, with proper parameters, this relaxation will not change the solution, which generally comes from the properties of the linear loss.

One attractive property for~\eqref{Zhang} over~\eqref{Plan} is that there is a closed-form solution for~\eqref{Zhang}, and thus, solving~\eqref{Zhang} is faster than~\eqref{Plan}, though both problems are equivalent in the sense that the solutions are the same for corresponding parameters $s$ and $\mu$. The convex penalty in~\eqref{Zhang} is replaced by several nonconvex penalties such as MCP~\cite{zhu2015towards} and $k$-support norm~\cite{chen_onebit_2015}. Better sampling complexities can be achieved for these nonconvex penalties and analytical solutions are obtained.

In this paper, we consider general penalties in~\eqref{Zhang} and derive efficient algorithms for many popular penalties by solving the dual problem. Even for many nonconvex penalties, our algorithms can find the global optimal solutions. These algorithms will help people investigate more theoretical properties for these nonconvex penalties such as the sampling complexity and consider better modifications such as adaptive sampling. %which are beyond the focus of this paper.

\section{Analytical Solutions for 1bit-CS} \label{sec:opt}

In this section, we consider the following generic optimization problem for robust 1bit-CS,
\begin{align}\label{pro:uncon}
\left.\begin{array}{ll}\Min\limits_{\vx\in\reals^n}~&\displaystyle f(\vx) - {1\over m}\sum\limits_{i=1}^my_i\langle\vu_i,\vx\rangle+{\tau\over2}\|\vx\|_2^2, \\
\textnormal{ subject~to }~&\|\vx\|_2^{2} \leq 1,
\end{array}\right.
\end{align}
where $\tau\geq 0$ and $f(\vx)$ is the penalty. Most existing papers assume that $\tau=0$, and the choice of $\tau>0$ is introduced by~\cite{zhu2015towards}. We will show that there is no need to choose a positive $\tau$ because optimal solutions do not depend on $\tau$ when $\tau$ is small enough and optimal solutions are not on the unit sphere for a large $\tau$. Let $\vv ={1\over m}\sum_{i=1}^my_i\vu_i$ and the objective function is
\begin{align*}
F(\vx)=f(\vx)-\langle \vv,\vx\rangle +{\tau\over2}\|\vx\|_2^2.
\end{align*}
We make the following assumption for $f(\vx)$.

\smallskip

\begin{assumption}
$f(\vx)\geq 0$ for all $\vx\in\reals^n$ and $f(\vzero)=0$.
\end{assumption}

\smallskip

The convexity of~\eqref{pro:uncon} depends on the penalty $f(\vx)$ and $\tau$.
For a convex $f(\vx)$, problem~\eqref{pro:uncon} is convex, and it could also be convex even if $f(\vx)$ is nonconvex when $\tau>0$ is large enough.
In order to find its global solution, we solve the corresponding dual problem and check whether the duality gap is zero, i.e., the optimal primal value is the same as the optimal dual value.
We define the corresponding Lagrangian functional as
\begin{align}
\mathcal{L}(\vx,\mu)= F(\vx)+{\mu\over 2} (\|\vx\|_2^2-1), \label{lagrangian}
\end{align}
and the following lemma gives a sufficient condition for a global optimal solution of problem~\eqref{pro:uncon}.

\smallskip
%%%%%   Revision  on 201704 %%%%%
\begin{lemma}{\cite[Theorem 6.2.5]{bazaraa_nonlinear_2006}}\label{lemma:optimal}
If there exist $(\vx^*,\mu^*)$ such that $\|\vx^*\|_2^2\leq 1$, $\mu^*\geq 0$, $\mathcal{L}(\vx^*,\mu^*)\leq \mathcal{L}(\vx,\mu^*)$ for all $\vx$, and $\mu^*(\|\vx^*\|_2^2-1)=0$, then
$\vx^*$ and $\mu^*$ are optimal solutions to the primal and dual problems, respectively, with no duality gap.
\end{lemma}

\smallskip

\begin{proof}
We have, for any $\mu\geq 0$,
\begin{align*}
\mathcal{L}(\vx^*,\mu) = & F(\vx^*)+{\mu\over 2} (\|\vx^*\|_2^2-1) \leq F(\vx^*) \\
= & F(\vx^*)+{\mu^*\over 2} (\|\vx^*\|_2^2-1)=\cL(\vx^*,\mu^*).
\end{align*}
{The inequality arises from the fact  $\|\vx^*\|_2^2\leq 1$, and the second equality holds because of $\mu^*(\|\vx^*\|_2^2-1)=0$.}
Thus, $(\vx^*,\mu^*)$ is a saddle point of $\cL(\vx,\mu)$, i.e.,
\begin{align}\label{eq:saddle_point}
\mathcal{L}(\vx^*,\mu) \leq \mathcal{L}(\vx^*,\mu^*)\leq \mathcal{L}(\vx,\mu^*),
\end{align}
for any $\mu\geq 0$ and $\vx$. Thus,
$$F(\vx^*)=\mathcal{L}(\vx^*,\mu^*)=\min_\vx\mathcal{L}(\vx,\mu^*).$$
The duality gap is zero, and $\vx^*$ and $\mu^*$ are optimal solutions to the primal and dual problems, respectively.
\end{proof}

\smallskip

Based on the lemma, we first solve the dual problem because it is concave and easy to solve in many cases. Then, we find $\vx^*$ and verify whether $\mu^*(\|\vx^*\|_2^2-1)=0$ is satisfied. If it is satisfied, then $\vx^*$ is a global optimal solution of~\eqref{pro:uncon}.

\subsection{Positive Homogeneous Penalties}
Assume that $f(\alpha\vx)=\alpha f(\vx)$ for any positive $\alpha$, i.e., $f(\vx)$ is positive homogeneous. We can obtain a global solution in two steps: a proximal operator and a normalization step. Some positive homogeneous penalties are listed here.
\begin{itemize}
\item $\ell_p$ norm ($0< p\leq +\infty$): e.g., $\ell_1$ norm~\cite{zhang2014efficient}; ~\footnote{$\ell_0$ ``norm'' is not positive homogeneous and can not be applied here.}
\item $\ell_p$ norm minus $\ell_q$ norm: e.g., $\ell_1-\ell_2$ norm~\cite{yin2015minimization,lou2017fast}.
\item $0$ function: Lemma 4.1 from paper~\cite{zhu2015towards}.
\item sorted $\ell_1$ penalty: nonconvex ones~\cite{huang2014nonconvex}; convex ones~\cite{bogdan2013statistical}; indicator function of $\ell_0$~\cite{chen_onebit_2015}; small magnitude penalized (SMAP)~\cite{zeng_decreasing_2014}.
\item One-sided norm~\cite{yang_onesided_2007}.
%\item elastic net:
\item Gauges~\cite{rockafellar_convex_1997}.
\end{itemize}

{Note that for a given  $\vx\neq 0$, we have $f(\vx)={d f(\alpha \vx)\over d\alpha} =\langle \tilde{\nabla} f(\alpha \vx),\vx\rangle$, where $\tilde\nabla f(\alpha\vx)$ is a generalized subgradient of $f$ at $\alpha\vx$~\cite[Definition 8.3]{rockafellar_variational_2004}. Let $\alpha=1$, and we have $f(\vx)=\langle \tilde{\nabla} f(\vx),\vx\rangle$.}
Define the proximal operator of $f$ as
$$\mbox{Prox}_f(\vv):=\argmin\limits_{\vx} f(\vx)+{1\over2}\|\vx-\vv\|_2^2,$$ and let
$\vt^*\in\mbox{Prox}_f(\vv)$. We have the following lemma.

\smallskip

\begin{lemma}\label{lemma1}
If $f(\vx)$ is positive homogeneous, we have that
$$f(\vt^*)-\langle \vt^*,\vv\rangle = -\|\vt^*\|_2^2.$$
\end{lemma}

\smallskip

\begin{proof}
When $\vt^*=\vzero$, we have $f(\vzero)=0$ because of $f(\vx)$ being positive homogeneous, and the result is trivial. When $\vt^*\neq \vzero$, we have $f(\vt^*)=\langle \tilde{\nabla} f(\vt^*),\vt^*\rangle$. Therefore,
\begin{align*}
f(\vt^*)-\langle \vt^*,\vv\rangle =& \langle \tilde\nabla f(\vt^*),\vt^* \rangle-\langle \vt^*,\vv\rangle\\
=&\langle \tilde\nabla f(\vt^*)-\vv,\vt^* \rangle=-\|\vt^*\|_2^2.
\end{align*}
The last equality is satisfied because $\vt^*\in\mbox{Prox}_f(\vv)$.
\end{proof}

\smallskip

\begin{theorem}If $f(\vx)$ is positive homogeneous and $\vt^*=\textnormal{Prox}_f(\vv)$, then an optimal solution for~\eqref{pro:uncon} is
\begin{align*}
\vx^*= \left\{\begin{array}{ll} {\vt^*/ \tau} (\mbox{or }\vzero\mbox{ if }\tau =0) & \textnormal{ if }\|\vt^*\|_2\leq \tau,\\
{\vt^*/ \|\vt^*\|_2} &\textnormal{ if }\|\vt^*\|_2>\tau.\end{array}\right.
\end{align*}
\end{theorem}

\smallskip

\begin{proof}
When $\tau>0$, we have
\begin{align*}
&\mathcal{L}(\vx,\mu) = f(\vx)+{\tau+\mu\over2}\left\|\vx-{\vv\over \tau+\mu}\right\|_2^2-{\|\vv\|_2^2\over 2(\tau+\mu)}-{\mu\over2}.
\end{align*}
Then $\vx^* = \vt^*/(\tau+\mu)$ is optimal for a given $\mu$, and
\begin{align*}
\min_\vx\mathcal{L}(\vx,\mu) =& {2 f(\vt^*)+\|\vt^*\|_2^2-2\langle \vt^*,\vv\rangle\over 2(\tau+\mu)}-{\mu\over 2}
= {-\|\vt^*\|_2^2\over 2(\tau+\mu)}-{\mu\over 2}.
\end{align*}
The last equality comes from Lemma~\ref{lemma1}. Thus the dual problem is a concave function of $\mu$ and we can find $\mu^*$ as
\begin{align}
\mu^*=\left\{\begin{array}{ll} 0 & \textnormal{ if }\|\vt^*\|_2\leq \tau,\\
\|\vt^*\|_2-\tau &\textnormal{ if }\|\vt^*\|_2>\tau.\end{array}\right.
\end{align}
Therefore we have
\begin{align*}
\vx^*= \left\{\begin{array}{ll} {\vt^*/\tau} & \textnormal{ if }\|\vt^*\|_2\leq \tau,\\
{\vt^*/ \|\vt^*\|} &\textnormal{ if }\|\vt^*\|_2>\tau.\end{array}\right.
\end{align*}
Thus $\mu^*(\|\vx^*\|_2^2-1)=0$ is satisfied, and Lemma~\ref{lemma:optimal} shows that $\vx^*$ is a global optimal solution of~\eqref{pro:uncon}.

Let $\tau=0$. When $\mu>0$, we have
\begin{align*}
\min_\vx\mathcal{L}(\vx,\mu) = {-\|\vt^*\|_2^2\over 2\mu}-{\mu\over 2} <0.
\end{align*}
Then we consider the case when $\mu=0$.
From Lemma~\ref{lemma1}, we have
\begin{align*}
\cL(\vt^*,0)=f(\vt^*)-\langle \vt^*,\vv\rangle =-\|\vt^*\|_2^2.
\end{align*}
Therefore, the positive homogeneity of $f$ gives
\begin{align*}
\min_\vx\mathcal{L}(\vx,0) = -\infty  \textnormal{ if } \vt^*\neq \vzero.
\end{align*}
When $\vt^*=0$, $\vt^*\in\mbox{Prox}_f(\vv)$ gives us that, for any $\vx$,
\begin{align*}
{1\over 2}\|\vv\|_2^2\leq f(\vx)+{1\over 2}\|\vx-\vv\|_2^2,
\end{align*}
which implies
\begin{align*}
-{1\over 2}\|\vx\|^2\leq f(\vx)-\langle\vx,\vv\rangle,
\end{align*}
and the positive homogeneity of $f$ shows that $\cL(\vx,0)=f(\vx)-\langle\vx,\vzero\rangle\geq 0$ for all $\vx$ and $\cL(\vzero,0)=0$.
Therefore,
\begin{align*}
\min_\vx\mathcal{L}(\vx,0) = 0 \textnormal{ if } \vt^*= \vzero.
\end{align*}
Together, we have $\mu^*=\|\vt^*\|_2$ and $\vx^*=\vt^*/\mu=\vt^*/\|\vt^*\|_2$ if $\vt^*\neq\vzero$.
When $\vt^*=\vzero$, we have $\mu^*=0$ and $\vx^*=\vzero$.

When $\tau=0$, Lemma~\ref{lemma:optimal} tells us that
\begin{align*}
\vx^*= \left\{\begin{array}{ll} \vzero & \textnormal{ if }\|\vt^*\|_2= 0,\\
{\vt^*/ \|\vt^*\|_2} &\textnormal{ if }\|\vt^*\|_2>0,\end{array}\right.
\end{align*}
is a global optimal solution of~\eqref{pro:uncon}. In fact, there may be multiple global solutions when $\vt^*=\vzero$ (See~\cite{huang_pinball_2015} for examples).
\end{proof}

In sum, a globally optimal solution of~\eqref{pro:uncon} can be obtained in two steps: a proximal operator and a normalization step.
\begin{algorithm}[!ht]
   \caption{General Algorithm for Positive Homogeneous Penalties}
   \begin{algorithmic}
   \STATE {\bfseries Input:} $\vv$, $f$
     \STATE {\bfseries Output:} $\vx$
		\STATE $\vt^*=\textnormal{Prox}_f(\vv)$
		\STATE $\vx^*={\vt^*/ \|\vt^*\|_2}$
%   \UNTIL{$noChange$ is $true$}
\end{algorithmic}
\end{algorithm}

\begin{remark}
When $\tau < \|\vt^*\|_2$, we have $\vx^*=\vt^*/\|\vt^*\|_2$, i.e.,  $\vx^*$ does not depend on $\tau$. When $\tau > \|\vt^*\|_2$, we have $\vx^*=\vt^*/\tau$, i.e.,  $\vx^*$ is not on the unit sphere. Therefore, there is no need to choose a positive $\tau$, and we let $\tau=0$ in the numerical experiments. {Note this result is consistent with Lemma 4.1 of~\cite{zhu2015towards} which shows that oracle estimators will not change when $\tau$ is small enough. }
\end{remark}

\subsection{General Penalties}
For a general $f(\vx)$, we consider the dual function
\begin{align}\label{eqn:Gdef}
G(\mu)=\min_\vx\mathcal{L}(\vx,\mu).
\end{align}
Given $\mu$, let $\vx^*(\mu)$ be an optimal solution of~\eqref{eqn:Gdef} defined as
\begin{align}
\vx^*(\mu)\in&\argmin_\vx\cL(\vx,\mu)
=\argmin_\vx f(\vx)+{(\tau+\mu)\over2}\left\|\vx-{\vv\over \tau+\mu}\right\|_2^2. \label{eqn:proxmial_f}
\end{align}
The following theorem provides the subdifferential of $G$.

\smallskip

\begin{theorem}\label{thm:subgradient}
Given $\mu\geq 0$, we have, for any $\tilde\mu\geq0$,
\begin{align}
G(\tilde\mu)\leq G(\mu)+{1\over2}(\|\vx^*(\mu)\|^2-1)(\tilde\mu-\mu).
\end{align}
\end{theorem}

\smallskip

\begin{proof}
{Using the definition of $G$ in~\eqref{eqn:Gdef}, we derive
\begin{align*}
G(\tilde\mu)=&\min_\vx\cL(\vx,\tilde\mu)  \leq \cL(\vx^*(\mu),\tilde\mu)\\
%=&F(\vx^*(\mu)) +{\tilde\mu\over 2}(\|\vx^*(\mu)\|^2-1)\\
%=&F(\vx^*(\mu)) +{\mu\over 2}(\|\vx^*(\mu)\|^2-1)\\
=&\mathcal{L}(\vx^*(\mu),\mu)+{1\over2}(\|\vx^*(\mu)\|_2^2-1)(\tilde\mu-\mu)\\
=&G(\mu)+{1\over2}(\|\vx^*(\mu)\|_2^2-1)(\tilde\mu-\mu),
\end{align*}
where the second equality follows from the definition of $\cL$ in~\eqref{lagrangian}, and the last equality is valid because of~\eqref{eqn:Gdef} and~\eqref{eqn:proxmial_f}.}
\end{proof}

\smallskip

The previous theorem shows that $(1-\|\vx^*(\mu)\|_2^2)/2\in\partial (-G)(\mu)$, where $\partial (-G)$ is the subdifferential of $-G$.
Note that when there are multiple optimal solutions of~\eqref{eqn:Gdef} for a given $\mu$, the subdifferential of $-G$ is $[\min \{(1-\|\vx^*(\mu)\|_2^2)/2\},\max \{(1-\|\vx^*(\mu)\|_2^2)/2\}]$.
Then $G(\mu)$ being concave gives us a way to find the optimal $\mu^*$.
{For general penalties, we turn to solve the dual problem, i.e., finding the maximizer of $G(\mu)$. The dual function is concave and has one variable, so many optimization methods can be applied.
However, in the evaluation of the subgradient of $G$, a proximal operator is needed. Therefore, many evaluations of the proximal operator is needed.}

When $\mu=+\infty$, we have $\vx^*(+\infty)=\vzero$.
If there exists an optimal solution of~\eqref{eqn:Gdef} such that $\|\vx^*(0)\|_2^2<1$, then $G(\mu)$ is decreasing for $\mu\in[0,+\infty)$, and we have $\|\vx^*(\mu)\|_2^2<1$ for all $\mu>0$ and the optimal $\mu^*=0$. Then $\vx^*(0)$ is an optimal solution of~\eqref{pro:uncon} because of Lemma~\ref{lemma:optimal}.
Otherwise, we have to find $\mu^*$ such that $\|\vx^*(\mu^*)\|_2^2=1$ or $0\in \partial (-G)(\mu^*)$.
If we find $\mu^*$ such that $\|\vx^*(\mu^*)\|_2^2=1$ is satisfied, then $\vx^*(\mu^*)$ is an optimal solution of~\eqref{pro:uncon} by Lemma~\ref{lemma:optimal}, otherwise $\mu^*(\|\vx^*\|_2^2-1)=0$ is not satisfied and whether $\vx^*(\mu^*)$ is an optimal solution of~\eqref{pro:uncon} is unknown.

\begin{remark}The following example shows that $\vx^*(\mu^*)$ can still be optimal for~\eqref{pro:uncon} even when $\mu^*(\|\vx^*\|_2^2-1)=0$ is not satisfied for the optimal $\mu^*$.
Let $F(x)=\|x\|_0-x/2$, then we have that the optimal solution is $x^*=0$.
The dual function of $\mu$ is
$$G(\mu)=\min\left(-{\mu\over2},1-{1\over 8\mu}-{\mu\over2}\right),$$
and the optimal $\mu^*=1/8$. The optimal $x^*$'s for $\mu^*$ are $0$ and $4$. Thus, we can still find $x^*(\mu^*)=0$ as a global optimal solution of $F(x)$.
However, in this case, the primal-dual gap is not zero.
\end{remark}

\begin{remark}
If $\tau$ is small enough such that $\|\vx^*(0)\|_2>1$ for all $\vx^*(0)$,  $\vx^*$ does not depend on $\tau$. When $\tau$ is large enough such that $\|\vx^*(0)\|_2<1$ for some $\vx^*(0)$, then $\vx^*$ is not on the unit sphere. Therefore, there is no need to choose a positive $\tau$, which is the same as in the case of positive homogeneous penalties, and we let $\tau=0$ in the numerical experiments.
\end{remark}

In order to find a global optimal solution of~\eqref{pro:uncon} efficiently, we have to make sure that the proximal operator has an analytical solution, because the proximal operator is evaluated for multiple times. Some penalties that have analytical solutions are:
\begin{itemize}
\item MCP and its generalizations~\cite{zhang2010nearly,Parekh2016enhanced,selesnick2017sparse},
\item SCAD~\cite{fan2001variable},
\item $\ell_0$ norm,
\item $\ell_{1/2}$ regularization~\cite{xu_LL12_2012},
%\item generalized shrinkage by Chartrand
\item Partial regularization~\cite{lu_sparse_2015}.
%\item Partly quadratic penalty~\cite{zhang2010nearly,Parekh2016enhanced}
\end{itemize}
In the following subsections, we describe algorithms for MCP, $\ell_0$ norm, and the nonconvex sorted $\ell_1$. Our algorithm is different from that in~\cite{zhu2015towards} for MCP. %We let $\tau=0$ because of

\subsection{Minimax Concave Penalty}
Let $f(\vx)=\sum_{i=1}^ng_{\lambda,b}(x_i)$ and $g_{\lambda,b}$ be defined as
\begin{align*}
g_{\lambda,b}(x)=\left\{\begin{array}{ll}\lambda |x|-{x^2/(2b)}, &\textnormal{ if }|x|\leq b\lambda,\\
{b\lambda^2/ 2}, &\textnormal{ if }|x|>b\lambda,\end{array}\right.
\end{align*}
for fixed parameters $\lambda>0$ and $b>0$. The analytical solutions for~\eqref{eqn:proxmial_f} can be obtained.

When $\mu\leq {1/b}$, we have
\begin{align}\label{eq:mcp_1}
x^*(\mu)= \left\{\begin{array}{ll}0, & \textnormal{ if }{v^2}\leq b\lambda^2\mu,\\
{v/ \mu},&\textnormal{ if }{v^2}\geq b\lambda^2\mu,\end{array}\right.
\end{align}
and when $\mu>1/b$, we have
\begin{align}\label{eq:mcp_2}
x^*(\mu)= \left\{\begin{array}{ll}0, & \textnormal{ if }{|v|}\leq \lambda,\\
{|v|-\lambda\over \mu-{1\over b}}\sgn(v), & \textnormal{ if }\lambda < |v|\leq b\lambda\mu,\\
{v/ \mu},&\textnormal{ if }{|v|}\geq b\lambda\mu.\end{array}\right.
\end{align}
For some $\mu$, we have two optimal solutions, as shown in the formulation.
The resulting algorithm is shown in Alg.~\ref{alg:MCP}.

\begin{algorithm}[!ht]
   \caption{MCP}
   \label{alg:MCP}
\begin{algorithmic}
   \STATE {\bfseries Input:} $\lambda$, $b$
     \STATE {\bfseries Output:} $\mu$
     \STATE Initialize: $\mu=1/b$
   \STATE $v_{[1]},v_{[2]},\dots,v_{[n]}=\mbox{ Sort}(|v_1|,|v_2|,\dots,|v_n|)$
     \STATE Find $L$ such that $v_{[L]}\leq \lambda< v_{[L+1]}$
     \STATE $d_2= \sum_{i=L+1}^n v_{[i]}^2$
     \STATE $d_{\max} = b^2d_2$
     \IF {$d_{\max}>1$}
     \STATE $i=L+1$; $d_1=0$
     \WHILE {$d_{\max}>1$}
     \STATE $\mu = {v_{[i]}}/(b\lambda)$; $d_{\max}=d_1/(\mu-1/b)^2+d_2/\mu^2$
         \IF {$d_{\max}<1$}
     \STATE Solve  $d_1/(\mu-1/b)^2+d_2/\mu^2=1$; return
         \ENDIF
     \STATE $d_1 = d_1+ (v_{[i]}-\lambda)^2$; $d_2=d_2-v_{[i]}^2$
     \STATE $d_{\max} = d_1/(\mu-1/b)^2+d_2/\mu^2$
     \STATE $i=i+1$
     \ENDWHILE
%     \STATE Solve  $d_1/(\mu-1/b)^2+d_2/\mu^2=1$
     \ELSE
     \STATE $i= L$;
     \WHILE {$d_{\max}< 1$}
     \STATE $\mu = v_{[i]}^2/(b\lambda^2)$; $d_{\max} = d_2/\mu^2$
         \IF {$d_{\max}>1$}
         \STATE $\mu = \sqrt{d_2}$; return
         \ENDIF
         \STATE $d_2 = d_2+v_{[i]}^2$
     \STATE $d_{\max} = d_2/\mu^2$
         \STATE $i=i-1$
     \ENDWHILE
%     \STATE $\mu = \sqrt{d_2}$
     \ENDIF
%   \UNTIL{$noChange$ is $true$}
\end{algorithmic}
\end{algorithm}

\subsection{$\ell_0$ norm}
Let $f(\vx)=\lambda\|\vx\|_0$, and the analytical solutions for~\eqref{eqn:proxmial_f} is:
\begin{align*}
x^*(\mu)= \left\{\begin{array}{ll}0, & \textnormal{ if }{v^2}\leq 2\lambda\mu,\\
{v/ \mu},&\textnormal{ if }{v^2}\geq 2\lambda\mu.\end{array}\right.
\end{align*}
The resulting algorithm is shown in Alg.~\ref{alg:L0}.

\begin{algorithm}[!ht]
   \caption{$\ell_0$ norm}
   \label{alg:L0}
\begin{algorithmic}
   \STATE {\bfseries Input:} $\lambda$
     \STATE {\bfseries Output:} $\mu$
     \STATE Initialize: $i=n$
%   \REPEAT
   \STATE $v_{[1]},v_{[2]},\dots,v_{[n]}=\mbox{ Sort}(|v_1|,|v_2|,\dots,|v_n|)$
     \STATE $\mu = {v_{[i]}^2/(2\lambda)}$; $d=v_{[i]}^2$; $d_{\max}= d/\mu^2$
   \WHILE{$d_{\max}<1$}
     \STATE $i=i-1$; $\mu = v_{[i]}^2/(2\lambda)$; $d_{\max}=d/\mu^2$
     \IF {$d_{\max} >1$ }
     \STATE $\mu=\sqrt{d}$; return
     \ELSE
   \STATE $d=d+v_{[i]}^2$; $d_{\max} = d/\mu^2$
     \ENDIF
   \ENDWHILE
%   \UNTIL{$noChange$ is $true$}
\end{algorithmic}
\end{algorithm}

\subsection{Sorted $\ell_1$ Penalty}

Let $f(\vx) = \lambda \sum_{i=1}^n w_{i} \left|x_{[i]}\right|$, where
\[
x_{[1]},x_{[2]},\dots,x_{[n]}=\mbox{Sort}(|x_1|,|x_2|,\dots,|x_n|)
\]
is sorted by the absolute component values.
Since weight $w_{i}$'s are assigned according to the sort, this regularization is called sorted $\ell_1$ penalty.
When $w_{1} \leq w_{2} \leq \ldots \leq w_{n}$, it is convex~\cite{bogdan2013statistical}. Otherwise, it is nonconvex and can be used to enhance sparsity~\cite{huang2014nonconvex}. For the nonconvex case, a typical weight setting is
\begin{equation}\label{weight_SL1}
w_{i} = \left\{
\begin{array}{ll}
1, & i < n_1, \\
\exp(-5 i/n_1), & i \geq n_1,
\end{array}
\right.
\end{equation}
where $n_1$ is a parameter related to the sparsity. Since the signal in 1bit-CS is very sparse,  $n_1 = 10$ is used in the numerical experiments.
An optimal solution can be analytically given by
\begin{align*}
t_{[i]}^*= &  \max \left\{|v_{[i]}| - w_{i} \lambda,0\right\}\sgn(v_{[i]}).
\end{align*}
Because this penalty is positive homogeneous, we apply the proximal operator first and then a normalization step.
The corresponding algorithm is given in Alg.~\ref{alg:SL1}.

\begin{algorithm}[!ht]
   \caption{Sorted $\ell_1$ Penalty}
   \label{alg:SL1}
\begin{algorithmic}
   \STATE {\bfseries Input:} $\lambda$, $\vw$ (decreasing)
     \STATE {\bfseries Output:} $\mu$
     \STATE Initialize: $\mu=0$
     \STATE $v_{[1]},v_{[2]},\dots,v_{[n]}=\mbox{Sort}(|v_1|,|v_2|,\dots,|v_n|)$
%   \REPEAT
   \FOR {$i = 1 : n$}
     \STATE $t_{[i]} = \max\{|v_{[i]}| - w_{i} \lambda,0\}\sgn(v_{[i]})$
   \ENDFOR
     \IF {$\|\vt\|>0$}
     \STATE $\mu= \|\vt\|$
     \ENDIF
%   \UNTIL{$noChange$ is $true$}
\end{algorithmic}
\end{algorithm}

\section{Numerical Experiments}\label{sec:num}

In numerical experiments, we randomly choose $K$ components from a $n$-dimensional signal, draw their values from the Gaussian distribution, and normalize the signal onto the unit $\ell_2$-norm ball.
Then, $m$ sign observations are generated by~\eqref{noise-data}, where $\varepsilon$ is the Gaussian noise with noise level $s_n$, which stands for the ratio between the variances of the measurements and $\varepsilon$.
We also consider sign flips with ratio 10\%. All the experiments are done with Matlab 2014b on Core i5-3.10GHz and 8.0GB RAM.

Before considering the recovery accuracy, we compare the computational time of Alg.~\ref{alg:MCP} and the algorithm in~\cite{zhu2015towards}. Both algorithms solve the same problem with the MCP penalty.
By the concavity of the dual function and Theorem~\ref{thm:subgradient}, the dual function is piecewise smooth and its subgradient is decreasing.
So we can find the optimal $\mu^*$ or the interval that contains the optimal $\mu^*$, and we find $\vx^*$ from~\eqref{eq:mcp_1} or~\eqref{eq:mcp_2}.
Therefore, there is at most one single variate problem, i.e., $d_1/(\mu-1/b)^2+d_2/\mu^2=1$, to solve.
While in~\cite{zhu2015towards}, this problem is solved for $n-L$ times, and there are many redundant computation steps.

To numerically compare the computational time, several pairs of $m$ and $n$ are considered.
For a fair comparison, we choose the same parameters for the MCP regularization $g_{\lambda,b}(x_i)$ as $\lambda = 0.1$ and $b = 3$.
Then the average and the standard derivation of computational times over $100$ trials are reported in Table~\ref{table-time}, where computational time of the Passive algorithm~\cite{zhang2014efficient} is given as well.

\begin{table}[htbp]

\begin{center}
\begin{small}
\begin{sc}
\caption{Average Computational Time}\label{table-time}
\vskip 0.3cm
\begin{tabular}{rr|rrr}
\toprule
%\abovespace\belowspace
$m$  & $n$ & Passive~~~ & Zhu's Alg.~~~~ & Alg. \ref{alg:MCP}~~~~ \\
\midrule
$500$  & $1000$  & $3.6 \pm 1.2~~\mathrm{ms}$  &  $1.51 \pm 0.04~~\mathrm{s}$~~ & $4.0 \pm 0.8~~\mathrm{ms}$\\
$1000$ & $1000$  & $6.7 \pm 1.4~~\mathrm{ms}$  &  $1.76 \pm 0.05~~\mathrm{s}$~~ & $8.9 \pm 1.2~~\mathrm{ms}$ \\
$1000$ & $2000$  & $14.8 \pm 1.4~~\mathrm{ms}$ &  $6.95 \pm 0.14~~\mathrm{s}$~~ & $18.5 \pm 1.3~~\mathrm{ms}$ \\
$2000$ & $2000$  & $25.1 \pm 1.7~~\mathrm{ms}$ &  $7.15 \pm 0.16~~\mathrm{s}$~~ & $30.2 \pm 2.2~~\mathrm{ms}$\\
$5000$ & $5000$ &  $148 \pm 5.1~~\mathrm{ms}$  & $43.6 \pm 1.48~~\mathrm{s}$~~  & $184 \pm 17~~\mathrm{ms}$ \\
\bottomrule
\end{tabular}
\end{sc}
\end{small}
\end{center}
\vskip -0.1in
\end{table}

The above result illustrates that the proposed analytical solution based algorithm can significantly reduce the computational burden from Zhu's algorithm.
{Compared with the Passive algorithm, which solves the $\ell_1$ minimization problem, Alg.~\ref{alg:MCP}, which solves the nonconvex MCP regularized problem, but the difference in computational time is minor. The comparison in performance is in the rest of this section.}
%takes a longer time but the recovery quality can be improved, as theoretically analyzed and numerically confirmed by~\cite{zhu2015towards}.

As discussed previously, our analysis covers many possible nonconvex regularizations. Section~\ref{sec:opt} gives several such kinds of algorithms including
\begin{itemize}
\item Alg.~\ref{alg:MCP} for minimizing the MCP penalty;
\item Alg.~\ref{alg:L0} for $\ell_0$ minimization;
%\item Algorithm \ref{alg:generalized} for generalized shrinkage;
\item Alg.~\ref{alg:SL1} for the nonconvex sorted $\ell_1$ penalty.
\end{itemize}
It can be expected that those nonconvex regularizers can improve the recovery quality from the $\ell_1$ norm when there are not plenty of measurements. In the following, we will report the performance of Alg.~\ref{alg:MCP}-\ref{alg:SL1} and the passive algorithm. %But since the comparison between MCP and the $\ell_1$ minimization has been fully considered by~\cite{zhu2015towards}, our experiments focus on the three nonconvex regularizations.

To select the parameters, we consider the following two method: i) choose the parameters based on the $\ell_2$ distance to the real signal, which results in ``ideal'' parameters. With ideal parameters, we can evaluate the best performance of each algorithm in each data sets; ii) tune the parameters by cross-validation based on consistency, which is a practical method. The selected parameters are not necessarily the best, especially when there are only a few observations. The comparison between the ideal and the selected parameters also implies the robustness of these methods to different parameters.

\begin{figure}[htb]
  \centering
  \subfigure[]{
    \label{fig-m:a} %% label for first subfigure
    \includegraphics[width=0.4\linewidth]{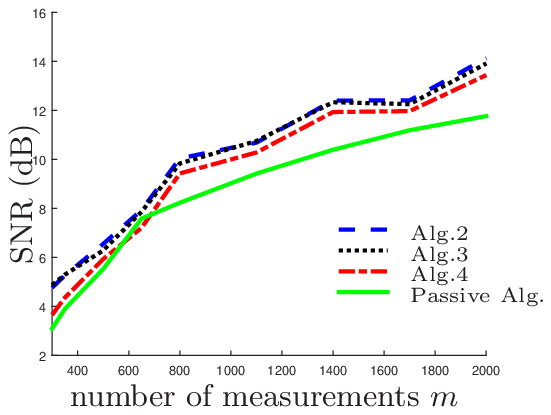}}\;\;
  \subfigure[]{
    \label{fig-m:b} %% label for first subfigure
    \includegraphics[width=0.4\linewidth]{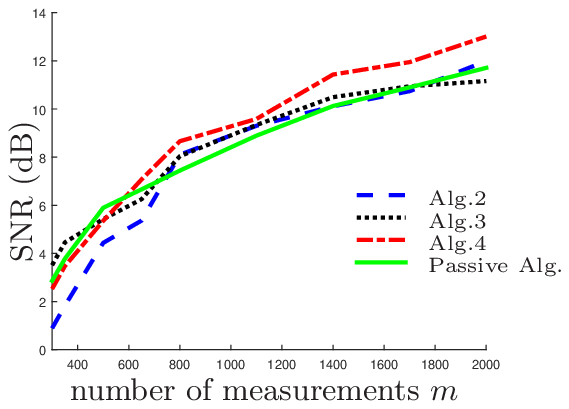}}
  \caption{Recovery performance for different numbers of measurements: MCP minimization (blue dashed line), $\ell_0$ minimization (black dotted line), sorted $\ell_1$ penalty (red dash-dotted line), and $\ell_1$ minimization (green solid line). In this experiment $n = 1000, K = 15, s_n = 10$, and sign flip ratio is $10\%$. (a) using the ideal parameters; (b) using parameters selected by 10-fold cross-validation.}\label{fig-m}
\end{figure}

First, we vary the number of measurements $m$ from $300$ to $2000$ and report the recovered quality in Fig.~\ref{fig-m}, where the signal-to-noise ratio in dB, i.e.,
\[
\mathrm{SNR}_\mathrm{dB}(\bar \vx, \tilde \vx) = 10 \log_{10} \left( \frac{\|\bar \vx\|^2_{2}}{\|\bar \vx - \tilde \vx\|^2_{2}} \right),
\]
is used to measure the quality of the recovered signal ($\bar \vx$ is the real signal and $\tilde \vx$ is the recovered one). Unless the amount of measurements is too small that no meaningful recovered signal can be obtained, the three nonconvex penalties can improve the performance from $\ell_1$ minimization if the optimal parameters can be obtained. In this case, MCP and $\ell_0$ achieve high recovery quality.
The SNRs obtained by the sorted $\ell_1$ minimization is a bit worse. When we select parameters by cross-validation, Alg.~\ref{alg:SL1} performs better than Alg.~\ref{alg:MCP} and~\ref{alg:L0}. The comparison indicates that the sorted $\ell_1$ is more stable to different parameters.

Notice that the four algorithms have no significant difference on computational time and, among these three nonconvex methods, Alg.\ref{alg:SL1} is most efficient. Specifically in this experiment, when $n = 1000, m = 1000$, the average computational times are: 6.7ms (Passive), 8.9ms (Alg.\ref{alg:MCP}), 8.2ms (Alg.\ref{alg:L0}), and 6.8ms (Alg.\ref{alg:SL1}). When $n = 10000, m = 5000$, the average computational times are: 284ms (Passive), 301ms (Alg.\ref{alg:MCP}), 295ms (Alg.\ref{alg:L0}), 291ms (Alg.\ref{alg:SL1}).

\begin{figure}[htb]
  \centering
  \subfigure[]{
    \label{fig-sn:a} %% label for first subfigure
    \includegraphics[width=0.45\linewidth]{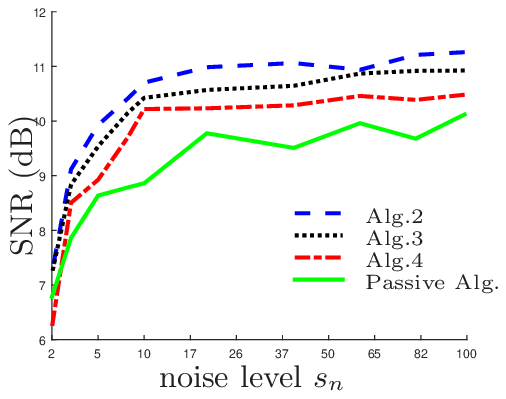}}\;\;
  \subfigure[]{
    \label{fig-sn:b} %% label for first subfigure
    \includegraphics[width=0.45\linewidth]{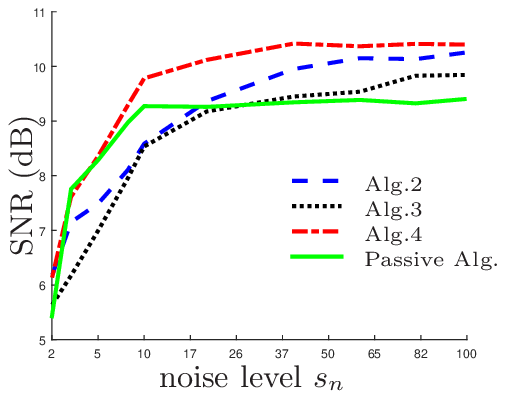}}
  \caption{Recovery performance for different noise levels: MCP minimization (blue dashed line), $\ell_0$ minimization (black dotted line), sorted $\ell_1$ penalty (red dash-dotted line), and $\ell_1$ minimization (green solid line). In this experiment $n = 1000, K = 15, m = 1000$, and sign flip ratio is $10\%$. (a) using the ideal parameters; (b) using parameters selected by 10-fold cross-validation.}\label{fig-sn}
\end{figure}

Similar observations can be found in Fig.~\ref{fig-sn}, where different noise levels are considered.
Generally, the three algorithms can both tolerate the existence of noise and outliers ($10\%$ of the sign measurements are flipped).
When the noise is not heavy, e.g., when ratio between the variance of the noise and that of the real measurements is below $0.1$, the three proposed algorithms have good noise suppression.

\begin{figure}[htb]
  \centering
  \subfigure[]{
    \label{fig-k:a}%% label for first subfigure
    \includegraphics[width=0.45\linewidth]{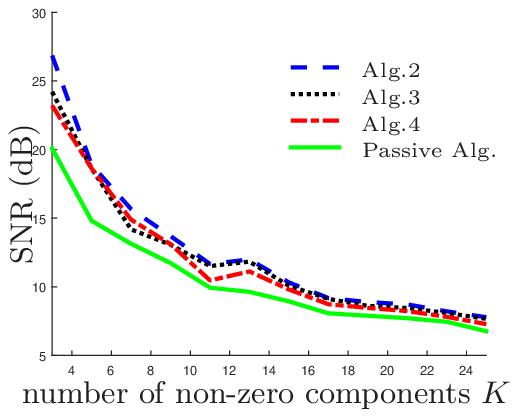}}\;\;
  \subfigure[]{
    \label{fig-k:b} %% label for first subfigure
    \includegraphics[width=0.45\linewidth]{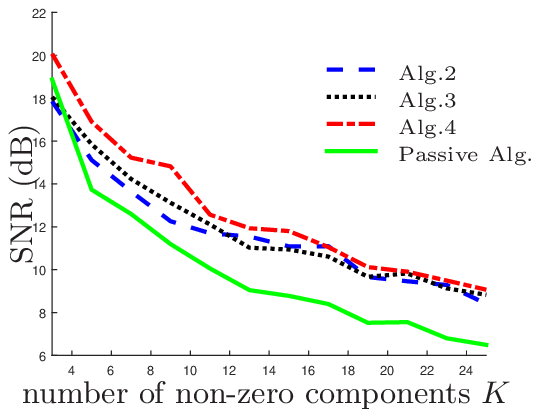}}
  \caption{Recovery performance for different sparsity levels: MCP minimization (blue dashed line), $\ell_0$ minimization (black dotted line), sorted $\ell_1$ penalty (red dash-dotted line), and $\ell_1$ minimization (green solid line). In this experiment $n = 1000, m = 1000, s_n = 10$, and sign flip ratio is $10\%$. (a) using the ideal parameters; (b) using parameters selected by 10-fold cross-validation.}\label{fig-k}
\end{figure}

Last, we consider different numbers of non-zero components $K$ with $n = 1000,~m = 1000,$ and $s_n =10$.
For the sorted $\ell_1$ penalty, there is one parameter $n_1$ in its weight~\eqref{weight_SL1} that is related to the signal sparsity.
In the previous experiments where $K$ is fixed to be $15$, we use $n_1 = 10$ without tuning.
Though that value is not optimal, the performance of the sorted $\ell_1$ penalty is generally satisfying.
In this experiment, we select $n_1$ from $\{2, 4, ..., 16\}$ for different $K$'s.
Totally, there are two parameters to tune for the sorted $\ell_1$ penalty, the same as MCP.
In Fig.~\ref{fig-k:a} and~\ref{fig-k:b}, the average SNRs for ideal and selected parameters are displayed, respectively.

Besides SNR, there are also other signal recovery criteria including:
\begin{itemize}
\item angular error:
\[
\mathrm{AE}(\bar \vx, \tilde \vx) = \frac{1}{\pi} \arccos \left(\frac{\bar \vx^T \tilde \vx}{\|\tilde \vx\|_2}\right);
\]

\item inconsistency ratio:
\[
\mathrm{INR}(\bar \vx, \tilde \vx) = \frac{\left |\{i: \sgn(\vu_i^T \bar \vx) \neq \sgn(\vu_i^T \tilde \vx)\} \right|}{m};
\]

\item ratio of missing support:
\[
\mathrm{FNR}(\bar \vx, \tilde \vx) = \frac{|\mathrm{supp}(\bar \vx) \backslash \mathrm{supp}(\tilde \vx)|}{|\mathrm{supp}(\bar \vx)|},
\]
where $\mathrm{supp}(\vx)$ stands for the support set of $\vx$; in our numerical experiments, a component being non-zero means that its absolute value is larger than $10^{-3}$;

\item ratio of misidentified support:
\[
\mathrm{FPR}(\bar \vx, \tilde \vx) = \frac{|\mathrm{supp}(\tilde \vx) \backslash \mathrm{supp}(\bar \vx)|}{n - |\mathrm{supp}(\bar \vx)|}.
\]
\end{itemize}

\begin{figure}[htb]
  \centering
  \subfigure[]{
    \includegraphics[width=0.35\linewidth]{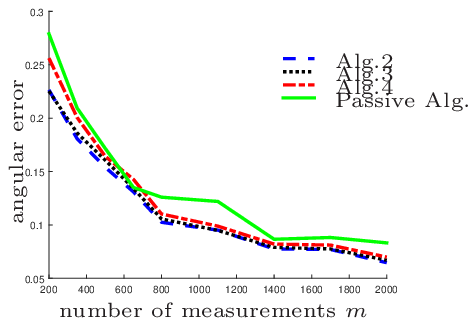}} \quad \quad
 \subfigure[]{
    \includegraphics[width=0.35\linewidth]{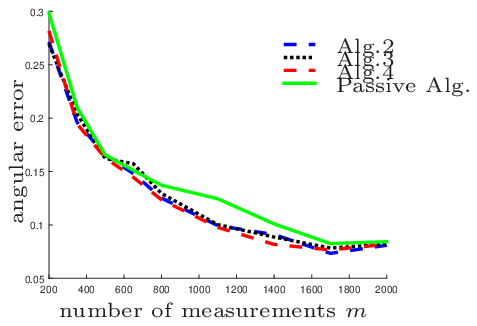}}\\
  \subfigure[]{
    \includegraphics[width=0.35\linewidth]{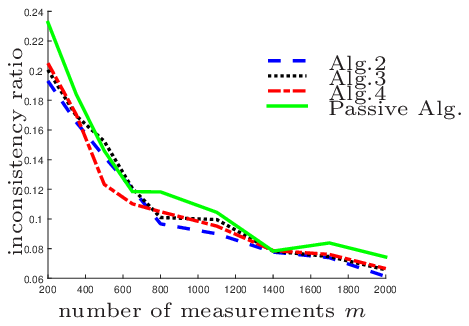}} \quad \quad
  \subfigure[]{
     \includegraphics[width=0.35\linewidth]{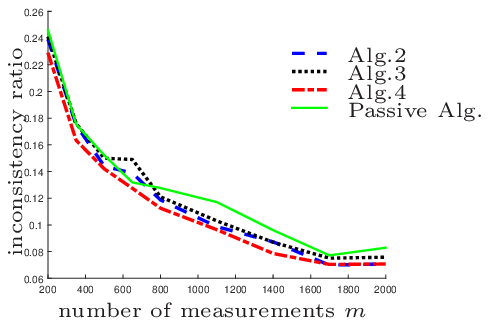}}\\
  \subfigure[]{
    \includegraphics[width=0.35\linewidth]{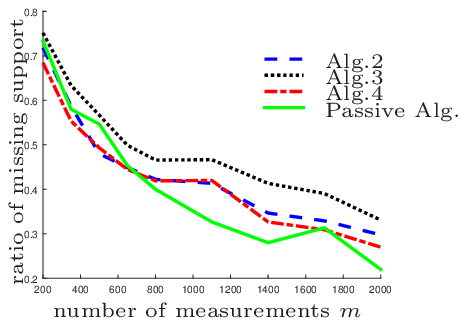}} \quad \quad
  \subfigure[]{
    \includegraphics[width=0.35\linewidth]{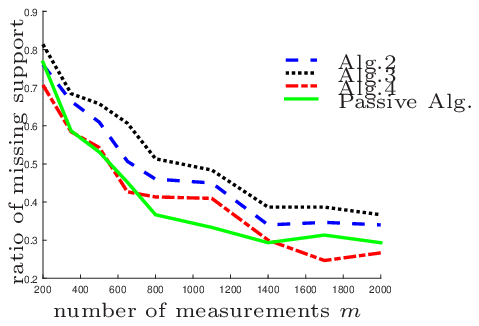}}\\
  \subfigure[]{
    \includegraphics[width=0.35\linewidth]{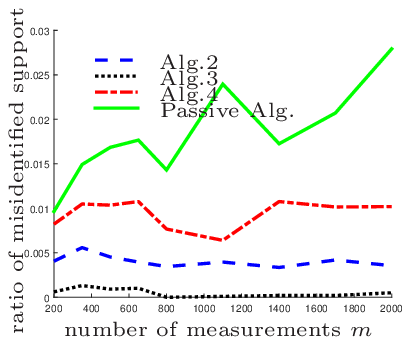}} \quad \quad
  \subfigure[]{
    \includegraphics[width=0.35\linewidth]{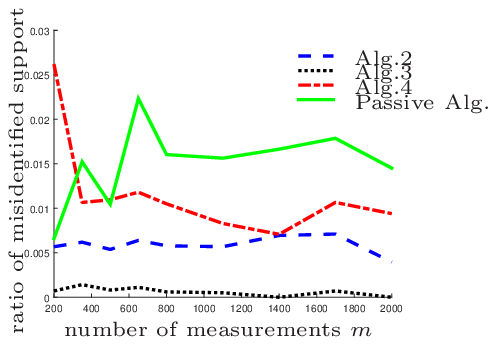}}
\caption{Recovery performance for different numbers of measurements. Three nonconvex regularizations are evaluated: MCP minimization (blue dashed line), $\ell_0$ minimization (black dotted line), sorted $\ell_1$ penalty (red dash-dotted line), and $\ell_1$ minimization (green solid line). In this experiment $n = 1000, K = 15, s_n = 10$, and sign flip ratio is $10\%$. Left column: use ideal parameters; Right column: use parameters selected by cross-validation. (a-b) angular error; (c-d) inconsistency ratio; (e-f) ratio of missing support; (g-h) ratio of misidentified support.}\label{fig-m-other}
\end{figure}

\begin{figure}[htb]
  \centering
  \subfigure[]{
    \includegraphics[width=0.35\linewidth]{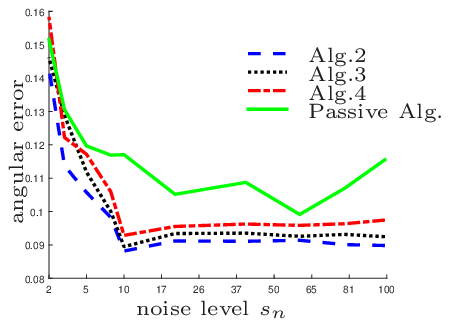}} \quad \quad
  \subfigure[]{
    \includegraphics[width=0.35\linewidth]{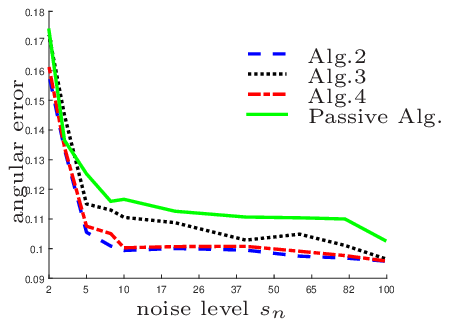}}\\
  \subfigure[]{
    \includegraphics[width=0.35\linewidth]{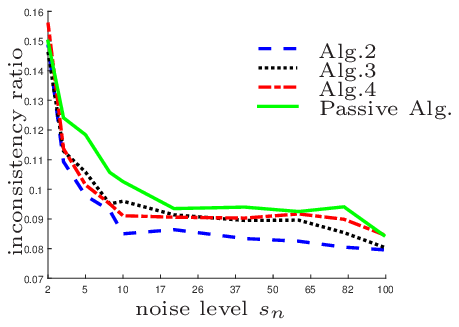}} \quad \quad
  \subfigure[]{
    \includegraphics[width=0.35\linewidth]{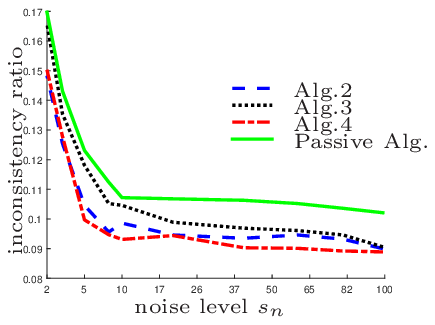}}\\
  \subfigure[]{
    \includegraphics[width=0.35\linewidth]{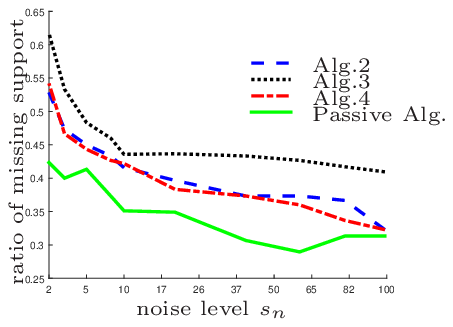}} \quad \quad
  \subfigure[]{
    \includegraphics[width=0.35\linewidth]{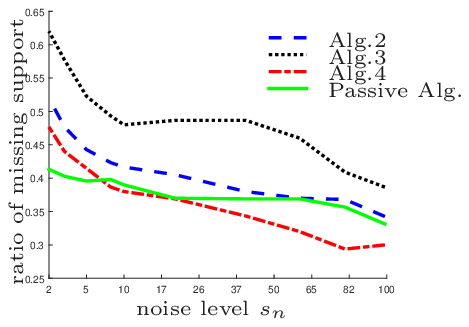}}\\
  \subfigure[]{
    \includegraphics[width=0.35\linewidth]{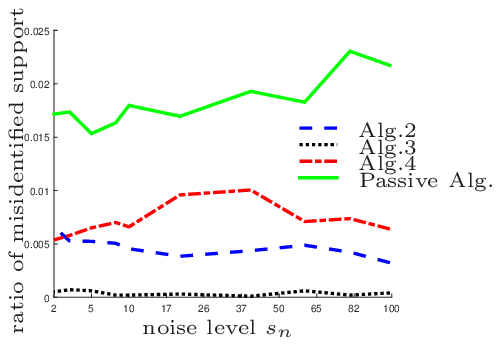}} \quad \quad
  \subfigure[]{
    \includegraphics[width=0.35\linewidth]{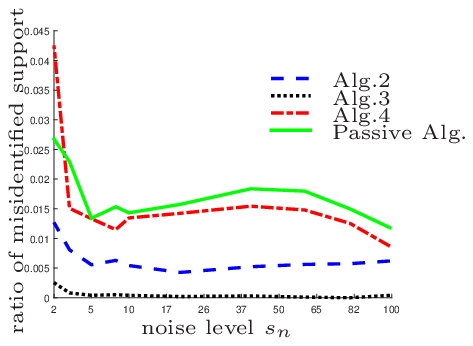}}
  \caption{Recovery performance for different noise levels. Three nonconvex regularizations are evaluated: MCP minimization (blue dashed line), $\ell_0$ minimization (black dotted line), sorted $\ell_1$ penalty (red dash-dotted line), and $\ell_1$ minimization (green solid line). In this experiment $n = 1000, K = 15, m = 1000$, and sign flip ratio is $10\%$. Left column: use ideal parameters; Right column: use parameters selected by cross-validation. (a-b) angular error; (c-d) inconsistency ratio; (e-f) ratio of missing support; (g-h) ratio of misidentified support.}\label{fig-sn-other}
\end{figure}

\begin{figure}[htb]
  \centering
  \subfigure[]{
    \includegraphics[width=0.35\linewidth]{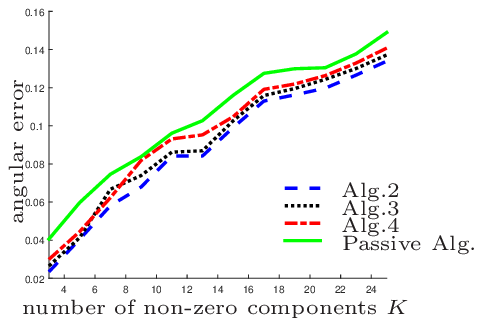}} \quad \quad
  \subfigure[]{
    \includegraphics[width=0.35\linewidth]{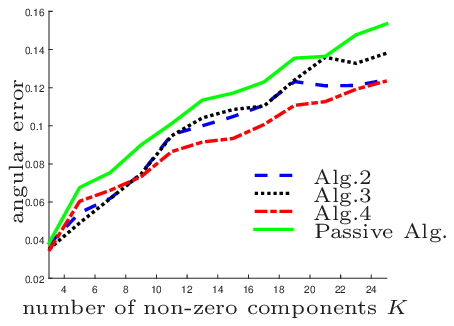}}\\
  \subfigure[]{
    \includegraphics[width=0.35\linewidth]{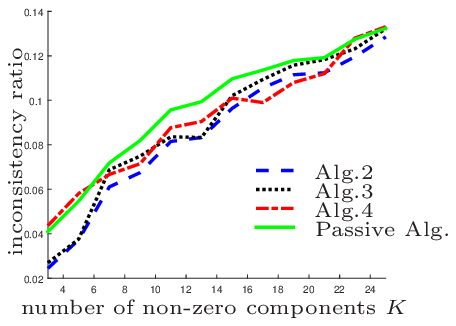}} \quad \quad
  \subfigure[]{
    \includegraphics[width=0.35\linewidth]{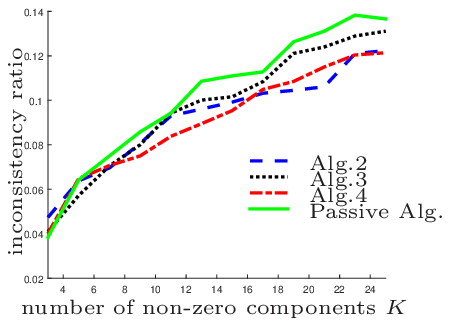}}\\
  \subfigure[]{
    \includegraphics[width=0.35\linewidth]{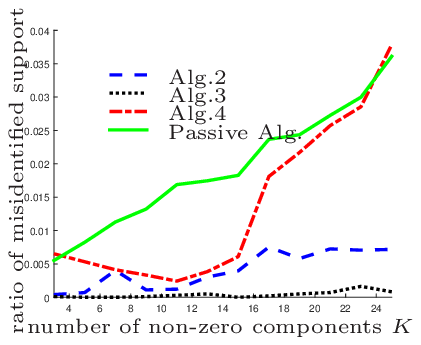}} \quad \quad
  \subfigure[]{
    \includegraphics[width=0.35\linewidth]{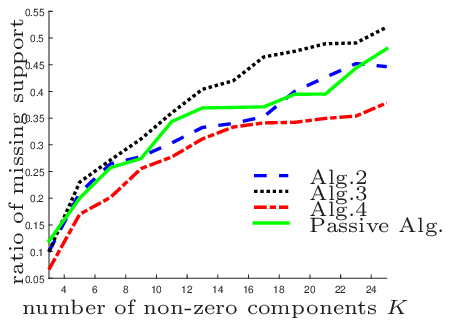}}\\
  \subfigure[]{
    \includegraphics[width=0.35\linewidth]{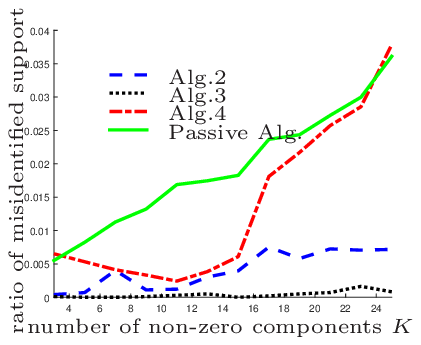}} \quad \quad
  \subfigure[]{
    \includegraphics[width=0.35\linewidth]{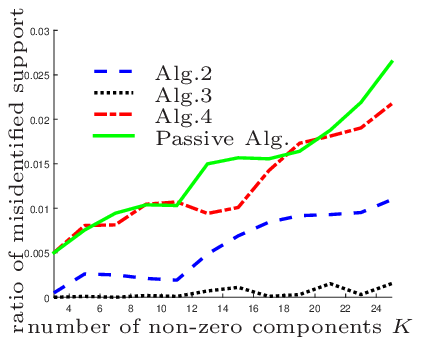}}
  \caption{Recovery performance for different sparsity levels. Three nonconvex regularizations are evaluated: MCP minimization (blue dashed line), $\ell_0$ minimization (black dotted line), sorted $\ell_1$ penalty (red dash-dotted line), and $\ell_1$ minimization (green solid line). In this experiment $n = 1000, m = 1000, s_n = 10$, and sign flip ratio is $10\%$. Left column: use ideal parameters; Right column: use parameters selected by cross-validation. (a-b) angular error; (c-d) inconsistency ratio; (e-f) ratio of missing support; (g-h) ratio of misidentified support.}\label{fig-K-other}
\end{figure}

To give multiple views for the considered nonconvex regularizations, we also report the performance measured by these criteria. The following figures are the average results of $100$ trials and the sub-figures (a-b), (c-d), (e-f), (g-h) correspond to AE, INR, FNR, and FPR, respectively. The performance for different numbers of measurements is reported in Fig.~\ref{fig-m-other}. Both the ideal and the selected parameters are used.  Similarly, the performance for different noise levels (Fig.~\ref{fig-sn-other}) and different sparsity levels (Fig.~\ref{fig-K-other}) are displayed. Together with SNRs shown before, we can have clear impression for the three proposed algorithms:
%In this section, we evaluated the three proposed algorithms, and the observations can be summarized as below:
\begin{itemize}
\item Alg.~\ref{alg:MCP} significantly reduces the computational time comparing to that given by~\cite{zhu2015towards} for MCP.
\item The efficiency of the proposed algorithms are comparable to the Passive algorithm~\cite{zhang2014efficient}, and the recovery quality is improved.
\item Alg.~\ref{alg:SL1} takes the least computational time.
Alg.~\ref{alg:MCP} and~\ref{alg:L0} have better performance when the parameters can be properly selected, otherwise, Alg.~\ref{alg:SL1} is better.
%Note that the performance is problem-dependent, so the comparison here may change for different data distributions and noise types.
\end{itemize}

\section{Conclusion}

Applying nonconvex regularizations is promising in enhancing the sparsity for 1bit-CS.
The major obstacle are that minimizing nonconvex regularizations usually requires long computational time and it is difficult to find the global optimal solution.
In this paper, we developed fast algorithms for several nonconvex regularizations based on its analytical solutions.
Our results extended the previous discussion on analytical solutions, which were limited to several specific regularizations, and also we significantly improved the computational efficiency for some problems. The proposed algorithms of several nonconvex regularizations are evaluated on numerical experiments and the computational time is comparable to the convex Passive algorithm, the currently fastest method for $\ell_1$ minimization of 1bit-CS. In the future, we will consider the nonconvex penalties in norm estimation~\cite{knudson2016one}, robust losses~\cite{mei2016landscape}, and adaptive thresholding~\cite{rusu2015adaptive, fang2016adaptive,baraniuk2017exponential}. These techniques are currently restricted to convex penalties, i.e., the $\ell_1$-norm minimization. It is promising to enhance the sparsity without introducing too much computational burden by applying the discussed analytical solutions.

\section*{Acknowledgment}

\small

This work was supported by National Science Foundation Grant DMS-1621798 and National Natural Science Foundation of China Grant 61603248.

The authors are grateful to the anonymous reviewers for their helpful comments.

\bibliographystyle{unsrt}        % Include this if you use bibtex
%\bibliography{ref_m1bit}

\end{document}